\documentclass[runningheads]{llncs}
\usepackage{graphicx}
\usepackage{comment}
\usepackage{amsmath,amssymb} % define this before the line numbering.

\usepackage{color}
\usepackage[dvipsnames]{xcolor}
\usepackage[american]{babel}
\usepackage[pagebackref=true,breaklinks=true,letterpaper=true,colorlinks,bookmarks=false]{hyperref}
\usepackage{floatrow}
\newfloatcommand{figurebox}{figure}[\nocapbeside][\dimexpr(\textwidth-\columnsep)/2\relax]
\newfloatcommand{tablebox}{table}[\nocapbeside][\dimexpr(\textwidth-\columnsep)/2\relax]
\usepackage[utf8]{inputenc}
\usepackage{babel}
\usepackage[font=small,labelfont=bf]{caption}

\newcommand{\wexu}[1]{\textcolor{black}{#1}}

 \newcommand{\jimmyref}[1]{\textcolor{black}{#1}}

\newcommand{\etal}{\textit{et al.}} 
\newcommand{\ie}{\textit{i.e.}}
\newcommand{\eg}{\textit{e.g.}}

\newcommand{\quotecap}[1]{``{\em #1}''}

\begin{document}
	\pagestyle{headings}
	\mainmatter
	\def\ECCVSubNumber{1417}  % Insert your submission number here

	\title{Compare and Reweight: Distinctive Image Captioning Using Similar Images Sets
	} % Replace with your title
	
	%	\begin{comment}
	\titlerunning{Distinctive Image Captioning Using Similar Images Sets}
%	\author{Jiuniu~Wang\inst{1,2,3}\orcidID{0000-0002-6113-0066} \and
%		Wenjia~Xu\inst{2,3}\orcidID{0000-0002-1425-4162} \and
%		Qingzhong~Wang\inst{2}\orcidID{0000-0003-1562-8098} \and
%		Antoni~B.~Chan \inst{2}\orcidID{0000-0002-2886-2513}}
	\author{Jiuniu~Wang\inst{1,2,3} \and
		Wenjia~Xu\inst{2,3,4}\and
		Qingzhong~Wang\inst{1} \and
		Antoni~B.~Chan \inst{1}}
	\authorrunning{J. Wang, W. Xu, Q. Wang, and A.B. Chan.}
	\institute{Department of Computer Science, City University of Hong Kong\\
		\email{\{jiuniwang2-c,qingzwang2-c\}@my.cityu.edu.hk, abchan@cityu.edu.hk} \and
	Aerospace Information Research Institute, Chinese Academy of Sciences \and
		University of Chinese Academy of Sciences\\
		\email{xuwenjia16@mails.ucas.ac.cn} \and
		Max Planck Institute for Informatics
		}
	%******************
	%******************
	\maketitle
	
    \begin{abstract}
         A wide range of image captioning models has been developed, achieving significant improvement based on popular metrics, such as BLEU, CIDEr, and SPICE. However, although the generated captions can accurately describe the image, they are generic for similar images and lack distinctiveness, \ie, cannot properly describe the uniqueness of each image. In this paper, we aim to improve the distinctiveness of image captions through training with sets of similar images. First, we propose a distinctiveness metric --- between-set CIDEr (CIDErBtw) to evaluate the distinctiveness of a caption with respect to those of similar images. Our metric shows that the human annotations of each image are not equivalent based on distinctiveness. Thus we propose several new training strategies to encourage the distinctiveness of the generated caption for each image, which are based on using CIDErBtw in a weighted loss function or as a reinforcement learning reward. Finally, extensive experiments are conducted, showing that our proposed approach significantly improves both distinctiveness (as measured by CIDErBtw and retrieval metrics) and accuracy (\eg, as measured by CIDEr) for a wide variety of image captioning baselines. These results are further confirmed through a user study. 
        Project page: \url{https://wenjiaxu.github.io/ciderbtw/}.
    \end{abstract}
	
	\section{Introduction}

    \begin{figure}
        \centering
        \includegraphics[height=5cm]{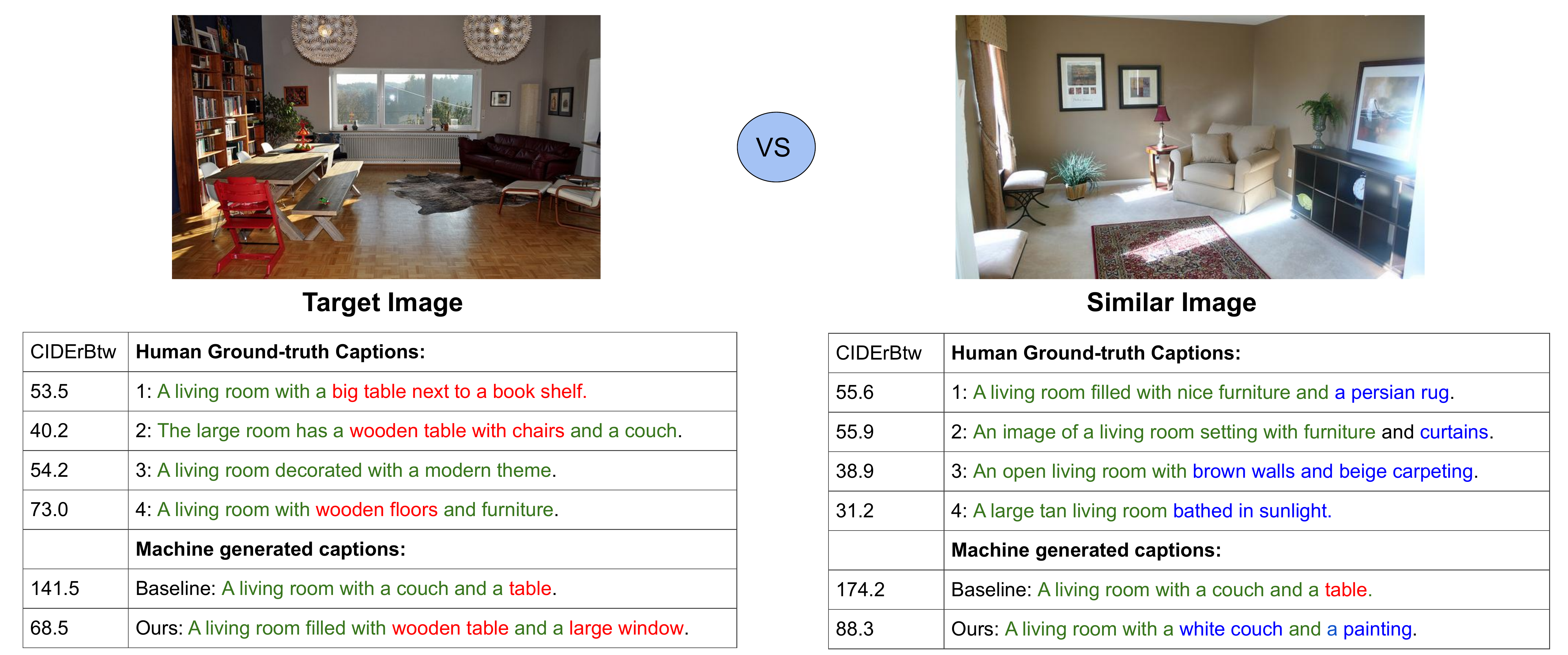}
        \caption{
        The human ground-truth captions of a target image and a semantically similar image contain both common words (highlighted in green) and distinctive words (highlighted in red for the target, and blue for the similar image).  The baseline model, Transformer~\cite{vaswani2017attention} trained with MLE and SCST, generates the same caption for both images, while our model generates distinctive captions with words unique to each image. The distinctiveness is measured using CIDErBtw, the CIDEr metric between the target caption and the GT captions of the similar images set, where lower values mean more distinctive.
        }
        \label{fig:teaser_figure}
    \end{figure}
	
Image captioning is attracting increasing attention from researchers in the fields of computer vision and natural language processing. It is promising in various applications such as human-computer interaction and medical image understanding~\cite{NIC,mrnn,templatemodel,spatt,convimagecap,cnnpluscnn}. Currently, the limitation of image captioning models is that the generated captions tend to consist of common words so that many images have similar or even the same captions (see Figure~\ref{fig:teaser_figure}). The distinctive concepts in images are ignored, which limits the application of image captioning. Although, auxiliary information such as where, when and who takes the picture could be used to generate personalized captions~\cite{chunseong2017attend,attend2u:2018:TPAMI}, many images do not have such information.
In terms of the quality of generated captions, \cite{liu2019generating} summarizes four attributes that encourage auto-generated captions to resemble human language: fluency, relevance, diversity, and descriptiveness. Various models and metrics have been proposed to improve the fluency and relevance of the captions so as to obtain accurate results. However, these captions are poor at mimicking the inherent characteristics of human language: \textit{distinctiveness}, which refers to the specific and detailed aspects of the image that distinguish it from other similar images.

Some recent works have focused on generating more diverse and descriptive captions, with techniques such as conditional generative adversarial networks (GANs)~\cite{cgan,cgan1}, self-retrieval~\cite{contrastive,luo2018discriminability,vered2019joint} and two-stage LSTM~\cite{liu2019generating}. Some works propose metrics for evaluating the {\em diversity} of a set of generated captions for a single image, 
based on the percentage of unique n-grams or novel sentences~\cite{cgan1} or the similarity between pairs of captions~\cite{my-div-paper}.
However, only encouraging the diversity, such as using synonyms or changing word order, may not help with generating distinctive captions among multiple similar images. For instance, the human caption in Figure~\ref{fig:teaser_figure} \quotecap{an image of a living room setting with furniture and curtains} is telling the same story as \quotecap{a living room with furniture and curtains}. 
Although the two sentences have different syntax and the first sentence is more diverse according to some metrics, the distinctiveness is not improved. 
In this paper, we mainly focus on promoting the \textit{distinctiveness} of image captioning, where the caption should describe the important and specific aspects of an image that can distinguish  it from other similar images. To evaluate distinctiveness,
the retrieval metric is generally employed in recent works~\cite{contrastive,liu2018show,luo2018discriminability,liu2019generating}.
However, using self-retrieval in captioning models could lead to repetition problem ~\cite{my-div-paper,vered2019joint}, \ie, the generated captions could repeat distinct words, which hurts language fluency.
Also, its result may vary when choosing different retrieval models or candidate images pool. 
In this work, we propose a general metric for distinctiveness, Between-Set CIDEr (CIDErBtw), by measuring the semantic distance between an image's caption and captions from a set of similar images. 
If the caption is distinct, i.e., captures unique concepts in its image, then it should have less overlap with its similar image set, i.e., lower CIDErBtw.
We found that the human annotations of each image are not equivalent based on distinctiveness. Consider the example image and caption pairs shown in 
Figure~\ref{fig:teaser_figure}, some ground-truth captions contain more distinct concepts (e.g., \textit{bathed in sunlight}) and detailed description that can distinguish the image from its similar image
(e.g., \textit{wooden floor} and \textit{brown walls}). However, traditional training objectives such as maximum likelihood estimation (MLE) and reinforcement learning (RL)  treat every ground-truth caption equally. 
Thus, one possible method for improving distinctiveness is to give more weight to the distinctive ground-truth captions during training.
In this way, the captioning model learns to focus on important visual objects or properties, and generate distinctive words instead of generic ones. 
	In summary, the contributions of our paper are three-fold:
    \begin{itemize}
        \item We propose a novel metric CIDErBtw to evaluate the distinctiveness of captions within similar image sets. Experiments show that our metric aligns with human judgment for distinctiveness.
        \item We use CIDErBtw as guidance for training, encouraging the model to learn from more distinctive captions. Experiments show that training with CIDErBtw is generic and yields consistent improvement for many baseline models. 
        \item  Based on the transformer network trained with SCST (self-critical sequence training)~\cite{rennie2017self} and CIDErBtw strategies, we generate distinctive captions while maintaining state-of-the-art performance according to evaluation metrics such as CIDEr and BLEU. Both automatic metrics and human evaluation demonstrate that our captions are more accurate and more distinctive. 
    \end{itemize}
	
	\section{Related work}
    \label{section:related_work}
    \noindent\textbf{Captioning models.} A wide range of image captioning models have been developed~\cite{NIC,mrnn,templatemodel,spatt,convimagecap,cnnpluscnn}, achieving satisfying results as measured by popular metrics, such as BLEU~\cite{bleu}, CIDEr~\cite{C} and SPICE~\cite{spice}. Generally, an image captioning model is composed of three modules: 1) visual feature extractor, 2) language generator,  and 3) the connection between vision and language. Convolutional neural networks (CNNs)~\cite{vggnet,he2016deep} are widely used as visual feature extractors. Recently, object-level features extracted by Faster-RCNN~\cite{ren2015faster} have also been introduced into captioning models~\cite{updown}, significantly improving the performance of image captioning models. \cite{yao2019hierarchy} proposed a hierarchy parsing model to fuse multi-level image features extracted by mask-RCNN~\cite{he2017mask}, which improves the performance of the baseline models. In terms of language generators, LSTMs~\cite{lstm1997} and its variants are the most popular, while some works~\cite{convimagecap,cnnpluscnn} use CNNs as the decoder since LSTMs cannot be trained in parallel. More recently, transformers~\cite{vaswani2017attention,radford2019language,devlin2018bert} show improved performance in both language generation and language understanding, where the multi-head attention plays the most important role and the receptive field is much larger than CNNs. Stacking multi-head attention layers could mitigate the long-term dependency problem in LSTMs. Hence, the transformer model could handle much longer texts. For vision-language connection, attention mechanisms~\cite{spatt,rennie2017self,updown,huang2019attention} are used to reveal the co-occurrence between concepts and objects in the images. 

    \noindent\textbf{Distinctive image captioning.} Previous works~\cite{contrastive,cgan,my-div-paper} reveal that training the captioning model with MLE loss or CIDEr reward result in over-generic captions, since the captioning models try to predict an ``average'' caption that is close to all ground-truth captions. These captions lack distinctiveness, \ie, they describe images with similar semantic content using the same caption. Recently, various works aim to solve this problem. In summary, they propose three aspects to consider: (1) {\em diversity}: describe one image with notably different expressions every time like humans~\cite{cgan}, or use rich and diverse wording~\cite{my-div-paper} to generate captions; (2) {\em discriminability}: describe an image by referring to the important and detailed aspects of the image, which is accurate, and informative~\cite{liu2018show,luo2018discriminability,liu2019generating,vered2019joint}; (3) {\em distinctiveness}: describe the important and specific aspects of an image that can distinguish the image from other similar images~\cite{contrastive,liu2019generating}. In our paper, we focus on the last aspect, distinctiveness.
    
    To promote diversity, some works~\cite{cgan,cgan1} employ GANs, where an evaluator distinguishes the generated captions from human annotations, encouraging the captions to be similar to human annotations. 
    Instead of using generative models, VisPara-Cap~\cite{liu2019generating} employs two-stage LSTM and visual paraphrases to improve diversity and discriminability, where the two-stage model is trained with a pair of ground-truth image captions from an image --- the first caption is less complex, and the next one with rich information is more distinctive.
    In contrast, our method is based on weighting all the ground-truth captions according to their distinctiveness, which retains more information for training.
    During inference,  VisPara-Cap~\cite{liu2019generating} first generates a simple caption and then paraphrases it into a more distinctive caption, which is a two-stage model and time-consuming. Another drawback of the model is that it cannot be trained in SCST~\cite{rennie2017self} manner, and therefore the performance based on BLEU~\cite{bleu}, CIDEr~\cite{C}, and SPICE~\cite{spice} is limited. In contrast, our method is able to improve both traditional metric scores and distinctiveness, and it can be applied to any image captioning model. 
    
    Contrastive learning~\cite{contrastive} and self-retrieval~\cite{liu2018show,luo2018discriminability,vered2019joint} are introduced into captioning models to improve the distinctiveness of the generated captions. 
    DiscCap~\cite{luo2018discriminability}, CL~\cite{contrastive} and PSST~\cite{vered2019joint} employ image retrieval to optimize the contrastive loss, which aims at pushing the generated caption far from other images in the training batch. On one hand, image retrieval encourages a model to generate distinctive words, while on the other hand, it hurts the accuracy and caption quality --- weighting too much on image retrieval could lead a model to repeat the distinctive words~\cite{my-div-paper}. 
    In contrast, we encourage the generated caption to learn from its own ground-truth captions, giving more weights to captions that are distinct from other similar images, and disregard those generic captions. Thus both accuracy and distinctiveness are promoted in our model.
    
    \noindent\textbf{Metrics for distinctiveness.} Traditional metrics such as BLEU~\cite{bleu}, METEOR~\cite{M}, ROUGE-L~\cite{R}, CIDEr~\cite{C} and SPICE~\cite{spice} normally consider the overlap between a generated caption and the ground-truth captions. 
    These metrics treat all ground-truth equally, and thus a generated caption that only uses common words could obtain high scores, reflecting the statistics of human annotations.
    Some works aim to generate multiple captions to cover more concepts in an image \cite{cgan,cgan1,posg,wang2019towards} and several diversity metrics are proposed, such as the number of novel captions, the number of distinct n-grams~\cite{Dist-K}, mBLEU~\cite{cgan1}, local and global word recall \cite{divmetric2}, and self-CIDEr~\cite{my-div-paper}. 
    However, these metrics only encourage the diversity and discriminability
    % of generated captions, 
    and do not explicitly evaluate distinctiveness. Although generating multiple captions could cover distinctive concepts, it is difficult to summarize them into one human-like description. 
    
	Currently, the retrieval approach is the most popular evaluation metric for distinctiveness\jimmyref{. A generated} caption is used as the query and a pre-trained image-text embedding model, \eg,  VSE++~\cite{faghri2017vse++}, is employed to rank the given images, with recall at $K$ (R@$K$) normally used to measure the distinctiveness of captions.
	Ideally, a correct and distinctive caption should retrieve the image that was used to generate the caption. The drawback of retrieval-based metrics is that they are time-consuming,  since it requires using a deep retrieval model. Moreover, different trained models could result in different R@$K$.
    In contrast, our proposed CIDErBtw metric for distinctiveness is fast and easy to implement, allowing it to be incorporated into various training protocols and captioning models.

	\section{Methodology}
    In this paper, we aim to obtain a distinct caption that describes the important, specific, and detailed aspects of an image. To achieve this goal, we train the captioning model to focus on important details that would distinguish the target image from semantically similar images. Our work involves two main components, the Between-Set CIDEr (CIDErBtw) that measures the distinctiveness of an image caption from those of similar images, and several strategies for training distinctive models based on CIDErBtw.
    
    The image captioning model aims to generate a sentence $c^*$ to describe the semantics of the target image $I_0$. 
    In the image caption dataset, the image $I_0$ is provided with $N$ annotated ground-truth captions $C^0 = \{c_1^0, c_2^0, \dots, c_N^0 \}$. We first find $K$ similar images $\{I_1,I_2,\dots,I_K\}$ that are semantically similar to $I_0$, and then calculate the CIDErBtw values of $C^0$ using these similar images. During training process, CIDErBtw can be used as an indicator of which ground-truth captions deserve more attention, or as a part of the reward in reinforcement learning (RL). This will train the model to generate %generating 
    a caption different from those of the similar images. Moreover, CIDErBtw can work as an evaluation metric to measure distinctiveness.

    \subsection{Similar images set}
    \label{section:similar_image_pair}
    
    According to the split of the training, validation, and testing dataset, we measure the similarity of the target image $I_0$ to every image within the same split. 
    For each image $I_0$ in the dataset, we find the top $K$ images $\{I_1, I_2, \dots, I_K\}$ with the highest semantic similarity to form a {\em similar images set}. 
    Similar images sets in the training split are used when calculating the loss and the reward during training, while those in the validation and test split are used to evaluate the distinctiveness of generated captions.
    
    Given every target image, we generate its similar images set according to an image-to-caption retrieval process. We \jimmyref{use VSE++~\cite{faghri2017vse++} to} encode images and captions into a joint semantic space, and obtain similar images sets via retrieval. 
	Given target image $I_0$, we obtain a set of closest captions $\{{c'_1}, {c'_2}, \dots, {c'_{N'}} \}$ in the joint space by image-to-caption retrieval, where $N'=N (K + 1)$ to ensure that at least $K$+1 images are obtained to construct the similar images set. The top $K$ images corresponding to this caption set are considered as similar to the target image $I_0$. 
	When using the retrieval method, the similarity of $I_i$ to $I_j$ denoted as ${S}({I_i},{I_j})$ can be expressed like
		\begin{align}
		{S}({I_i},{I_j}) &= \mathop{\max }\limits_{k\in\{1,\cdots,N\}} g_r(I_i, c_k^j),
		%k=1,2,\cdots,N \,, \\
		\quad g_r(I_i, c_k^j) = \frac{\phi (I_i)^T \theta (c_k^j)}{ \| \phi (I_i)  \|  \| \theta(c_k^j)  \|} \,,
		\end{align}
	where $g_r(I_i, c_k^j)$ represents the retrieval score between the target image $I_i$ and the $k$-th ground-truth caption of $I_j$, and  $\phi (\cdot)$ and $\theta (\cdot)$ are the image and caption encoders.

	\subsection{Between-set CIDEr (CIDErBtw)}
	Next, we introduce the definition of Between-set CIDEr (CIDErBtw) and its applications. In this paper, we mainly apply CIDErBtw in the following three aspects. During training, CIDErBtw is used to reweight the cross entropy (XE) loss and the \jimmyref{reinforcement learning (RL)} reward for each ground-truth caption. The CIDErBtw metric is also used directly as part of the reward to guide \jimmyref{RL}. During inference, CIDErBtw is used as a metric to measure the distinctiveness of a generated caption.
	
	\subsubsection{CIDErBtw definition.}
	
	CIDErBtw reflects the distinctiveness of a caption $c$ by measuring the similarity of $c$ to the captions of similar images \jimmyref{$C^{(s)}$}. Specifically,  given a caption $c$ for image $I_0$, the similar images set $\{I_1, I_2, \dots, I_K\}$ retrieved in Section~\ref{section:similar_image_pair} and their ground-truth captions $C^{(s)}=\{c^k_n\}_{n=1,k=1}^{N,K}$, we define the CIDErBtw score of $c$ as
	\begin{align}
	CIDErBtw(c) = \frac{1}{KN}\sum\limits_{k = 1}^{K}\sum\limits_{n = 1}^{N}{g_c(c,c^k_n)}, \label{equ:CIDErBtw}
	\end{align}
	where $N$ is the number of ground-truth captions provided for each image, $g_c(c,c^k_n)$ represents the CIDEr value between $c$ and $c^k_n$.  Actually, the methodology could be extended to use any caption metric to measure between-set similarity. Here we use CIDEr because it focuses more \jimmyref{on} low frequency words (through TF-IDF vectors) that could be more distinctive, is efficient to compute, and is the most frequently used metric to evaluate performance of image captioning models. 
	
	\subsubsection{CIDErBtw weight.}
    \label{sec: CIDErBtw weight}
    For conventional training strategies such as MLE and reinforcement learning, we maximize the likelihood or reward for the given ground-truth captions $C^0 = \{c_1^0, c_2^0, \dots, c_N^0 \}$. In previous methods, each ground-truth caption $c_i^0$ is treated equally, whereas these ground-truth might have different distinctiveness. In this work, we focus more attention to distinctive ground-truth captions by reweighting the training loss. 
    For every training image $I_0$, we provide its $N$ ground-truth captions $C^0$ with different weights $W=\{w_1,w_2,\dots,w_N\}$, according to their CIDErBtw scores $V=\{v_1,v_2,\dots,v_N\}$,
    \begin{align}
    {v_i} = CIDErBtw(c_i^0), \quad
    {w_i} = \lambda_w  - \alpha_w \frac{{{v_i}}}{{\mathop {\max }\limits_i ({v_i})}} , \label{eq:weight}
    \end{align}
    where $\lambda_w$ and $\alpha_w$ are hyperparameters. Here $w_i$ indicates the contribution of the $i$-th ground-truth caption during model training. More distinctive captions will have lower $v_i$, leading to higher weight $w_i$.

	\subsection{CIDErBtw training strategies}
	
	\begin{figure}[t]
		\centering
		\includegraphics[height=3.6cm]{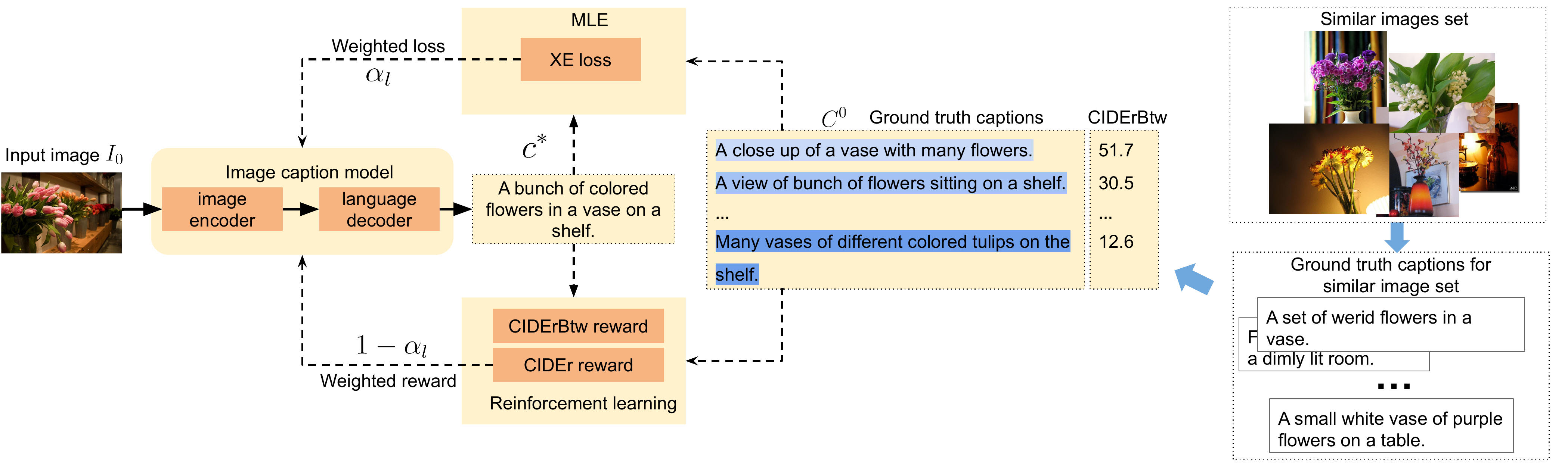}
		\caption{
		The framework of our CIDErBtw image captioning model. $\alpha_l$ is a hyperparameter that controls the weight of the two optimization modules. The solid and dashed lines represent the forward and backward process. $c^*$ and $C^0$ indicate the generated caption and the ground-truth captions. 
        With CIDErBtw, we reweight the ground-truth captions when calculating the XE loss and reward. The shade of blue shows the CIDErBtw weight $w_i$ for each caption.}
		\label{fig:structure}
	\end{figure}
	
	Figure~\ref{fig:structure} shows the overall framework of our CIDErBtw Image Caption model. The model is composed of a \jimmyref{image} encoder and \jimmyref{language} decoder. These two modules can generate a caption $c^*$ for input image $I_0$. There are two criteria to update the parameters of our image captioning model, the XE loss ${\cal L}_{XE}$ and RL reward ${\cal L}_{RL}$. We apply a hyperparameter $\alpha_l$ to control the weight of these two criteria,
	\begin{align}
	{\cal L} =  \alpha_l {\cal L}_{XE} + (1 - \alpha_l) {\cal L}_{RL}\,.
	\end{align}
	Following SCST (self-critical sequence training)~\cite{rennie2017self}, the training process of our model can be divided into two steps. The first step only trains with ${\cal L}_{XE}$, setting $\alpha_l=1$, and the second step only trains with ${\cal L}_{RL}$, setting $\alpha_l=0$.
	
	\subsubsection{Reweighting XE loss.}
	Given the words in a ground-truth caption $c_i^0 =\{d_1,d_2,\dots,d_T\}$, XE loss can be expressed as
	\begin{align}
	L_{XE}(c_i^0) =  - \sum\limits_{t = 1}^T \log  p_{\theta}({d_t}|d_{1:t - 1}, I_0),
	\end{align}
	where $p_{\theta}(d_t|d_{1:t-1}, I_0)$ denotes the probability of the word $d_t$ given the word sequence ${d_1},\dots,{d_{t - 1}}$ and image $I_0$. The CIDErBtw weighted XE loss is then 
	%can be expressed as
	\begin{align}
	{\cal L}_{XE} = \sum\limits_{i = 1}^{N} w_i L_{XE}(c_i^0)\,.
	\label{eqn:LXE}
	\end{align}

	\subsubsection{Reweighting RL reward.} 

	For RL, we reweight the CIDEr reward according to the CIDErBtw to focus more on distinctive captions, resulting in a new reward, 
	\begin{align}
	\tilde R(c^*) = \frac{1}{N}\sum\limits_{i = 1}^{N}w_i  g_c(c^*,c^0_i),
	\label{eqn:newreward}
	\end{align}
	where $g_c(c^*,c^0_i)$ is the CIDEr value between $c^*$ and ground-truth $c^0_i$.
	
	\subsubsection{CIDErBtw reward.}
	
	Finally, when performing RL, our CIDErBtw can also be used as a part of the reward related to distinctiveness.
	We combine the CIDErBtw score  with the prevous reward $\tilde R(c^*)$ and obtain the final RL reward $R(c^*)$ and RL loss ${\cal L}_{RL}$ as
	\begin{align}
	\label{equ:Reinforcement}
	R(c^*) = \tilde R(c^*) - \alpha_r CIDErBtw(c^*),  \quad
	{\cal L}_{RL} = - {\mathbb{E}_{{c^*} \sim  {p_\theta }}}[R(c^*)],
	\end{align}
	where $CIDErBtw(c^*)$ represents CIDErBtw score of the generated caption $c^*$ defined in (\ref{equ:CIDErBtw}), $\alpha_r$ is a hyperparameter controlling the relative contributions, and the greedy sampling is used as the RL policy $p_\theta$.

	\subsubsection{CIDErBtw evaluation metric.}
	
	CIDEr measures the similarity between the generated caption $c^*$ and its ground-truth captions $C^0$,
	and has become an important evaluation metric in image captioning.
	We believe the distinctiveness should also be measured when evaluating the quality of generated captions. Thus we propose to use CIDErBtw as a complementary evaluation metric for image captioning models.
	We hope that the caption $c^*$ generated by the model is closer to the semantics of target image $I_0$, while far from the semantics of other $K$ similar images $\{I_1, I_2, \dots, I_K\}$. Therefore, the $c^*$ generated by a more distinctive image captioning model will have a lower CIDErBtw.
% 	on the validation or test split.
	Note that for evaluation, the similar image sets are computed using the validation or test split.
% 	, where appropriate.
    \wexu{Note that CIDEtBtw requires human annotations to evaluate the generated captions, which is similar to other captioning evaluations, e.g., CIDEr~\cite{C}, BLEU~\cite{bleu}, METEOR~\cite{M}, and ROUGE~\cite{R}. Although VSE++ does not require human annotation for evaluation, it still needs ground-truth captions in the training phase, and the performance is highly related to the training data.}
	
	\section{Experiments}
    In this section, we conduct extensive experiments to evaluate the effectiveness of CIDErBtw in generating distinctive captions. Note that our motivation is to generate distinctive captions as well as achieve high caption quality.
    
    \subsection{Implementation details}
    \subsubsection{Dataset.}
    
    We use the MSCOCO dataset~\cite{mscocodataset} with Karpathy spliting~\cite{karpathy2015deep}. The numbers of images are 113,287 for training, 5,000 for validation, and 5,000 for testing. There are five annotated captions for each image.

    \subsubsection{Models.}
    For the image encoder, following Luo~\etal~\cite{luo2018discriminability}, we use two types of features in the experiments, \ie, the FC features and the spatial features. The FC features are extracted from Resnet-101~\cite{he2016deep}, and each image is encoded as a vector of dimension $2,048$. The spatial features are extracted from the output of a Faster-RCNN~\cite{ren2015faster} following UpDown~\cite{updown}. 
   
    Our experiments are performed using four baseline models, \ie, FC~\cite{rennie2017self}, Att2in~\cite{rennie2017self}, UpDown~\cite{updown}, and Transformer~\cite{vaswani2017attention}. FC model only uses the FC features, Att2in and Transformer  only use the spatial features, and UpDown  uses both types of features.
    Each model is trained using four methods: 1) MLE with standard XE loss, denoted as ``\textit{model}''; 
    2) MLE with CIDErBtw-weighted XE loss in (\ref{eqn:LXE}), denoted as ``\textit{model}+CIDErBtw'';
    3) SCST~\cite{rennie2017self}, which trains with standard XE loss first, and then switches to RL with CIDEr reward, denoted as ``\textit{model}+SCST'';
    4) SCST using weighted XE loss and weighted RL reward in (\ref{eqn:newreward}), denoted as ``\textit{model}+SCST+CIDErBtw''.

    \subsubsection{Training details.}
    % For different models, w
    We set $\lambda_w$ as $1.5$, $\alpha_w$ between $0.25$ to $1.25$ when reweighting the loss and the reward.
    $\alpha_r$ is set to $0.4$ when using CIDErBtw reward, and 0 otherwise. We use Adam~\cite{kingma2014adam} to optimize the training parameters with an initial learning rate $5\times 10^{-4}$ and a decay factor $0.8$ every three epochs. During test time, we apply beam search with size five to generate captions.

    \subsubsection{Metrics.}
    For evaluation we consider two groups of metrics. 
    The first group includes language quality metrics CIDEr, BLEU3, BLEU4, METEOR, ROUGE-L, and SPICE for evaluating the accuracy and quality of generated captions. The second group assesses the distinctiveness of captions, and includes our CIDErBtw metric and retrieval metrics (\ie, R@1, R@5, R@10). When calculating CIDErBtw, we collect $K = 5$ similar images for each target image, so the CIDErBtw score measures the similarity between the generated caption and $25$ captions from the similar images set, with lower values indicating more distinctiveness. Similar images sets are generated using a pre-trained VSE++~\cite{faghri2017vse++} to perform the caption-to-image retrieval (see Section \ref{section:similar_image_pair}). For the retrieval metrics, we follow the protocol in~\cite{liu2019generating,luo2018discriminability,contrastive}. Given a generated caption, images are retrieved in the joint semantic space of the pre-trained VSE++, with the goal to retrieve the original image. 
    Recall at $K$ (R@$K$) is used to measure the retrieval performance, where a higher recall represents a better distinctiveness.
    
    \subsection{Experiment results}
    In this section, we present the experiment results to show the effectiveness of CIDErBtw training strategies at improving caption distinctiveness.
    Due to space constraints, the ablation study is presented in the supplemental. 
    
    \subsubsection{Effect of CIDErBtw strategies.} 
    The main results are presented in the top and middle of Table~\ref{table:main_results}.  
    All baseline models obtain better performances when using CIDErBtw weighting in training process, for both MLE or SCST, which suggests that our method is widely applicable to many existing models. Specifically, our method both reduces the CIDErBtw score and improves other accuracy metrics, such as CIDEr.
    This shows that the generated captions become more similar to ground-truth captions, while more distinctive from other images' captions since redundancy is suppressed. 
    Among the four baseline models, CIDErBtw reweighted loss and reward have the largest effect on  Transformer~\cite{vaswani2017attention}. 
    Most likely the multi-head attention and larger receptive field of Transformer allow it better extract details and context from the image that is distinctive.
    	
	Next we apply all three of our CIDErBtw reward strategies together on Transformer+SCST, which is denoted as ``+CIDErBtwReward'' in Table~\ref{table:main_results}.  Compared to only using reweighted loss and reward (Transformer+SCST+CIDErBtw), adding the CIDErBtw reward in RL improves both the CIDErBtw and retrieval metrics significantly (i.e., improves distinctiveness), at the expense of a small decrease in accuracy (CIDEr).

    Finally, we examine the disadvantage of SCST that directly optimizing CIDEr reward improves the fluency of captions but also leads to common and generic words.
    Consistent with~\cite{my-div-paper,liu2019generating}, 
    the baseline models trained with SCST obtain higher CIDEr but also perform worse in CIDErBtw and R@$K$, compared with models trained only with MLE. 
    Optimizing the model with CIDErBtw weighted reward will relieve this problem, and the distinctness of captions will be promoted, while maintaining or even improving the overall quality of the captions. 
    
    \setlength{\tabcolsep}{2pt}

    \begin{table}[tb]
        \begin{center}
            \caption{
            Comparison of caption accuracy and distinctiveness on MSCOCO test split: (top) baseline models trained with MLE using standard or our weighted XE loss; (middle) models trained with SCST using standard or our weighted loss/reward; (bottom) SOTA methods for generating distinctive/discriminative captions.
            CIDEr, BLEU3/4, METEOR, ROUGE-L, and SPICE measure caption accuracy, while CIDErBtw and R@$K$ measure distinctiveness.
            $\uparrow$ or $\downarrow$ show whether  higher or lower scores are better for each metric. 
            CIDErBtw could not be computed for some models because the captions are not publicly available.
            Our self-retrieval results (R@$K$) and those of
            \cite{gu2018stack,luo2018discriminability,liu2019generating,contrastive} use the pre-trained VSE++ model and the same protocol. $\dagger$ Note that \cite{vered2019joint} reports self-retrieval results using a different retrieval model/protocol -- they use their own model for retrieval -- which makes it not directly comparable.
            }

			\label{table:main_results}
			\resizebox{340pt}{95pt}{
				\begin{tabular}{l|cc|ccccc|ccc}
					\hline \hline
					Method  & CIDEr$\uparrow$  & CIDErBtw$\downarrow$ & BLEU3$\uparrow$  & BLEU4$\uparrow$  & METEOR$\uparrow$ & ROUGE-L$\uparrow$ & SPICE$\uparrow$  & R@1$\uparrow$   & R@5 $\uparrow$  & R@10$\uparrow$  \\
					\hline\hline
FC~\cite{rennie2017self}                     & 97.90   & 83.35    & 41.81 & 31.58 & 25.22  & 53.34   & 17.99 & 15.44 & 40.36 & 55.08 \\
FC+CIDErBtw (ours)                                                 & 98.82  & 83.22    & 42.03 & 31.79 & 25.46  & 53.48   & 18.29 & 16.24 & 41.54 & 56.64 \\
	\hline
Att2in~\cite{rennie2017self}                 & 110.04 & 83.19    & 46.36 & 35.75 & 26.79  & 56.18   & 19.91 & 17.44 & 43.88 & 58.02 \\
Att2in+CIDErBtw (ours)                                             & 110.97 & 82.42    & 46.63 & 36.0    & 27.03  & 56.30    & 20.01 & 17.98 & 44.72 & 58.62 \\
	\hline

UpDown~\cite{updown}                         & 111.25 & 79.46    & 45.64 & 35.93 & 27.54  & 56.24   & 20.54 & 20.10 & 47.58 & 61.92 \\
UpDown+CIDErBtw (ours)                                             & 112.77 & 78.34    & 46.35 & 36.10  & 27.69  & 56.36   & 20.68 & 20.92 & 49.72 & 63.98 \\
	\hline

Transformer~\cite{vaswani2017attention}      & 110.13 & 80.98    & 44.80  & 34.46 & 26.98  & 55.30    & 20.18 & 21.52 & 49.88 & 64.70 \\
Transformer+CIDErBtw (ours)                                        & 112.44 & \textbf{75.35}    & 45.44 & 35.01 & 27.59  & 55.66   & 20.74 & 21.84 & 50.48 & 65.04 \\
	\hline
\hline
FC+SCST~\cite{rennie2017self}                & 104.43 & 90.09    & 43.10 & 31.59 & 25.46  & 54.33   & 18.67 & 11.44 & 33.16 & 48.04 \\
FC+SCST+CIDErBtw (ours)                                            & 104.76 & 89.41    & 43.25 & 31.72 & 25.60   & 54.35   & 18.58 & 11.74 & 33.62 & 48.32 \\
	\hline

Att2in+SCST~\cite{rennie2017self}            & 117.96 & 87.40     & 47.22 & 35.31 & 27.17  & 56.92   & 20.57 & 16.00 & 41.55 & 56.66 \\
Att2in+SCST+CIDErBtw (ours)                                        & 118.48 & 87.21    & 47.33 & 35.41 & 27.27  & 56.94   & 20.77 & 16.82 & 42.26 & 57.72 \\
	\hline

UpDown+SCST~\cite{updown}                    & 121.94 & 86.82    & 48.82 & 36.12 & 27.95  & 57.61   & 21.29 & 18.50 & 46.34 & 61.70 \\
UpDown+SCST+CIDErBtw (ours)                                        & 123.02 & 86.42    & 48.98 & 36.39 & 28.12  & 57.78   & 21.44 & 19.68 & 47.30 & 62.78 \\
	\hline

Transformer+SCST~\cite{vaswani2017attention} & 125.13 & 86.68    & 50.26 & 38.04 & 27.96  & 58.60    & 22.30  & 23.38 & 54.34 & 68.44 \\
Transformer+SCST+CIDErBtw (ours)                                   & \textbf{128.11} & 84.70     & \textbf{51.29} & \textbf{39.0}    & \textbf{29.12}  & \textbf{59.24}   & 22.92   & 24.46 & 55.22 & 69.02\\
\multicolumn{1}{r|}{+CIDErBtwReward (ours)}   & 127.78 & 82.74    & 50.97 & 38.52 & 29.09   & 58.82   & \textbf{22.96} & \textbf{26.46} & \textbf{57.98} & \textbf{71.28}\\

					\hline\hline
					Stack-Cap~\cite{gu2018stack} & 120.4          & 88.7     & 47.9 & 36.1   &   27.4    &  56.9         &   20.9       &    21.9       &    49.7      &    63.7       \\
DiscCap~\cite{luo2018discriminability} & 120.1          & 89.2        & 48.5 & 36.1   &    27.7     &57.8           & 21.4          &   21.6       &   50.3        & 65.4          \\
VisPara-Cap~\cite{liu2019generating} & 86.9          & -         & -  & 27.1      &  -    &-           & 21.1          &   26.3       &   57.2        & 70.8          \\
CL-Cap~\cite{contrastive}       & 114.2           & 81.3           & 46.0           & 35.3           & 27.1           & 55.9           & 19.7              & 24.1             & 52.5             & 67.5            \\
PSST~\cite{vered2019joint}  & 111.9           & -          & -             & 32.2           & 26.4           & 54.4           & 20.6           & 45.3$\dagger$              & 79.4$\dagger$              & 89.9$\dagger$              \\ % & 38.1*              & 74.2*              & 86.3*        
\hline \hline
				\end{tabular}
			}
		\end{center}
	\end{table}

    \subsubsection{Reasons for improving CIDEr.}
    Results in Table~\ref{table:main_results} show that models trained with CIDErBtw obtain better performance for {\em both} distinctiveness metrics and accuracy metrics. Given that our training method puts more weight on distinct ground-truth captions, it is expected that we will obtain lower CIDErBtw and higher R@$K$ scores. 
    However, the reason why our method also improves  caption accuracy (CIDEr) is less obvious, especially for SCST, which {\em directly} optimizes CIDEr using RL.
    Note that CIDEr is based on the cosine similarity between TFIDF vectors, and thus low-frequency words (with higher IDF weights) will have higher impact on the CIDEr score.  Since rare words are also distinct, their usage in a caption should increase CIDEr.
    If using distinct words can increase CIDEr, then why does RL with CIDEr reward not use distinct words? 
    We speculate that RL gets stuck in a local minimum of models that only use frequent words because of two reasons: 1) equal weighting of an image's ground-truth captions encourages the model to predict the common words that match all captions; and 2) regularization encourages models to use smaller vocabularies -- using less words means less non-zero weights in the network, and lower model complexity. 
    By reweighting the reward with CIDErBtw, more reward is obtained when using diverse words, which effectively moves the learning process out of this local minimum.

    \subsubsection{Comparison with state-of-the-art.}
    We list the performance of state-of-the-art captioning models that focus on distinctiveness at the bottom of Table~\ref{table:main_results}. 
    Compared to these models, our model (Transformer+SCST+CIDErBtw, and +CIDErBtwReward) generally achieves superior results in both accuracy and distinctiveness --- our model obtains both a high CIDEr score and low CIDErBtw score (or high retrieval score) at the same time.
    Specifically, Stack-Cap~\cite{gu2018stack} and DiscCap~\cite{luo2018discriminability} have lower accuracy (CIDEr 120) and less distinctiveness (CIDErBtw 89, R@1 22), compared to our model.
    VisPara-Cap~\cite{liu2019generating} has high distinctiveness by using visual paraphrases, slightly worse than our model (+CIDErBtwReward), while the accuracy (CIDEr 86.9) is much lower than our model.
    CL-Cap~\cite{contrastive} and PSST~\cite{vered2019joint} directly optimize the retrieval loss, aiming to identify the input image among a set of randomly-chosen distractor images, which improves the distinctiveness. 
    CL-Cap has similar distinctiveness as our method, obtaining worse R@$K$ than ours, but better CIDErBtw.\footnote{We could not compare distinctiveness with PSST since their captions are not publicly available, and they use a different evaluation protocol for R@$K$.}
    However, directly optimizing the training parameter with retrieval loss results in low-quality captions, lowering the accuracy (CIDEr 114.2 and 111.9) compared to our model.

	\subsection{User Study}
    To fairly evaluate the quality of generated sentences and verify the consistency between the metrics and human perspective, we conducted two user studies. Firstly, we performed a user study on image retrieval to assess distinctiveness, following the protocol in \cite{luo2018discriminability}. The task involves displaying the target image, \wexu{a semantically similar image which is retrieved following the method in Section~\ref{section:similar_image_pair},} and a generated caption describing the target image. The users are asked to choose the image that more closely matches the caption. 
    % The retrieval of
    % \cite{luo2018discriminability} retrieves the similar image based on Euclidean distances between visual features. However, images with close visual distance may convey different semantic meanings. Thus, we use similar images with close semantic distance to the target image (see Section~\ref{section:similar_image_pair}). 
    
    In the second experiment, we compare captions generated from a baseline model trained with and without CIDErBtw. In each trial, an image and two captions are displayed, and the user is asked to choose the better caption with respect to two criteria: distinctiveness and accuracy. 
    In each experiment, we randomly sample 50 similar images pair from the test split.
    We perform the experiment on four captioning models: UpDown~\cite{updown} and  Transformer \cite{vaswani2017attention} trained by SCST with and without CIDErBtw (denoted as UD,  UD+CIDErBtw, TF and TF+CIDErBtw).
    Twenty people participated in the user study, and we collected about $6,000$ responses in total. See the supplemental for more details.
	
	The results for the image retrieval user study are shown in Table~\ref{tab:user_study}. Compared to the baseline model, our method increases the accuracy of image retrieval by $5.6\%$ and $14.4\%$. 
	This user study is consistent with the automatic image retrieval results (R@$K$), and indicates that captions generated by our model are more distinctive in terms of both machine and human perception, than those of the baseline models.  

    The result for the distinctiveness/accuracy user study are shown in Figure~\ref{fig:user_study}. 
    From human perspective, captions from our models are more distinctive than the baseline models (our captions are selected 69\% and 72\% of the time). The improvement of accuracy is not as much (our model selected ~59\% and ~63\% of the time), since the baseline models already generate captions that are accurate. Again this is consistent with the observations from the machine-based metrics (CIDErBtw and CIDEr).

  \begin{minipage}[htb]{0.95\textwidth}
    \begin{minipage}[b]{0.47\textwidth}
    \centering
    \small
    \begin{tabular}{c|c}
    \hline 
    Method            & image retrieval   \\
    \hline
    UD            & 68.7\%  $^{**}$  \\
    UD+CIDErBtw & \textbf{74.3\%} $^{**}$\\
    \hline
    TF  & 75.2\% $^{*}$\\
    TF+CIDErBtw       & \textbf{79.6\%} $^{*}$        \\
    \hline 
    \end{tabular}
    \captionof{table}{User study on image retrieval to assess caption distinctiveness. Our models trained with CIDErBtw generated more distinctive captions, enabling the user to more accurately select the correct image, compared with the baselines (2-sample z-test on proportions,  $^*$ p$<$0.05, $^{**}$ p$<$0.01). }
    \label{tab:user_study}

    \end{minipage}
      \hfill
    \begin{minipage}[b]{0.47\textwidth}
    \centering
    \includegraphics[width=4.3 cm]{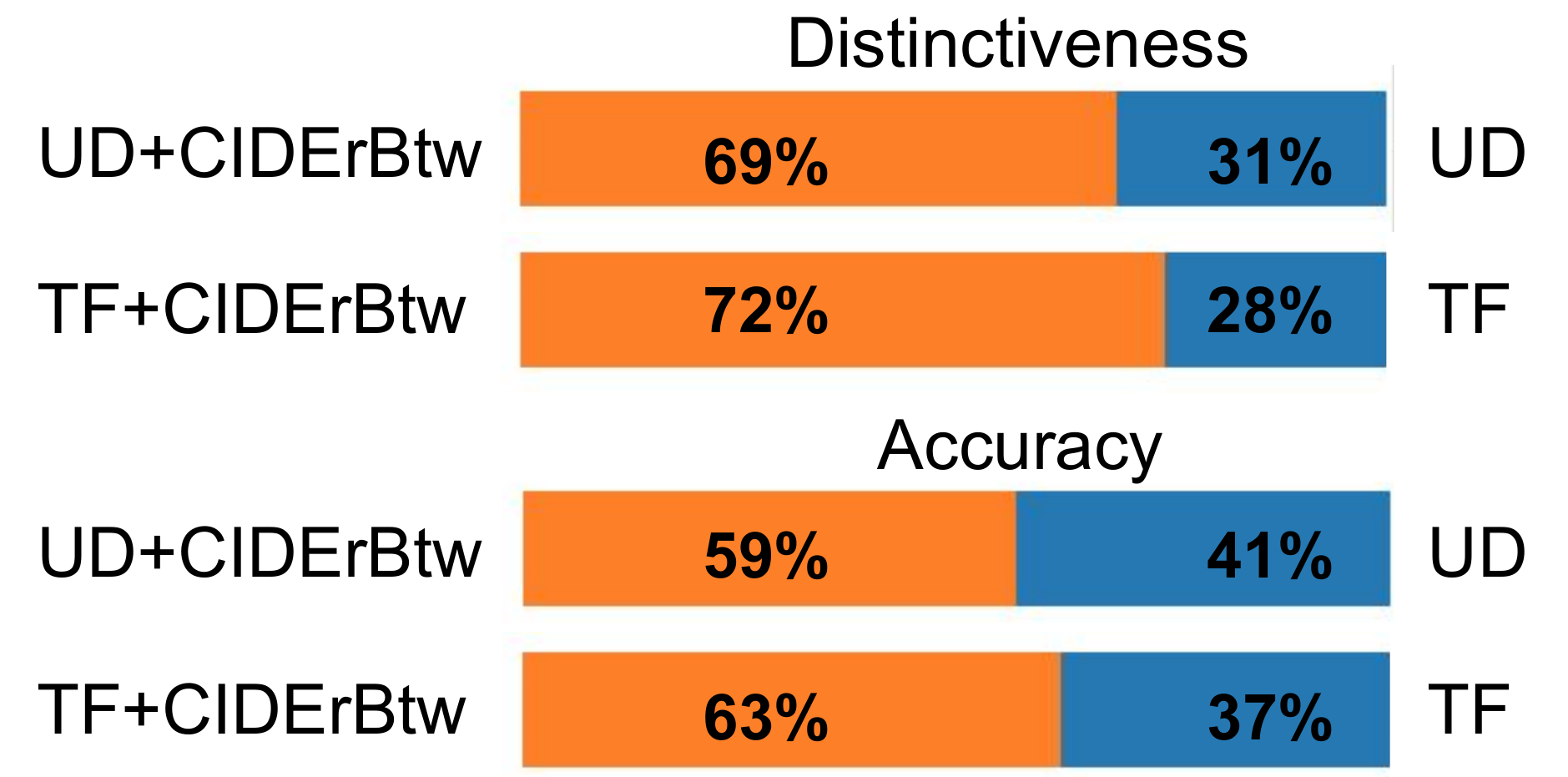}
    \captionof{figure}{User study comparing captions generated  from models trained with and without our CIDErBtw. Users selected our models trained with CIDErBtw more frequently when assessing accuracy and distinctiveness (Chi-Square test, p$<$0.001 for each pair).
    }
  \label{fig:user_study}
  \end{minipage}
  \end{minipage}

\subsection{Qualitative Results}
\label{sec:qualitative}
\begin{figure}[h!]
		\centering
		\includegraphics[width=12cm]{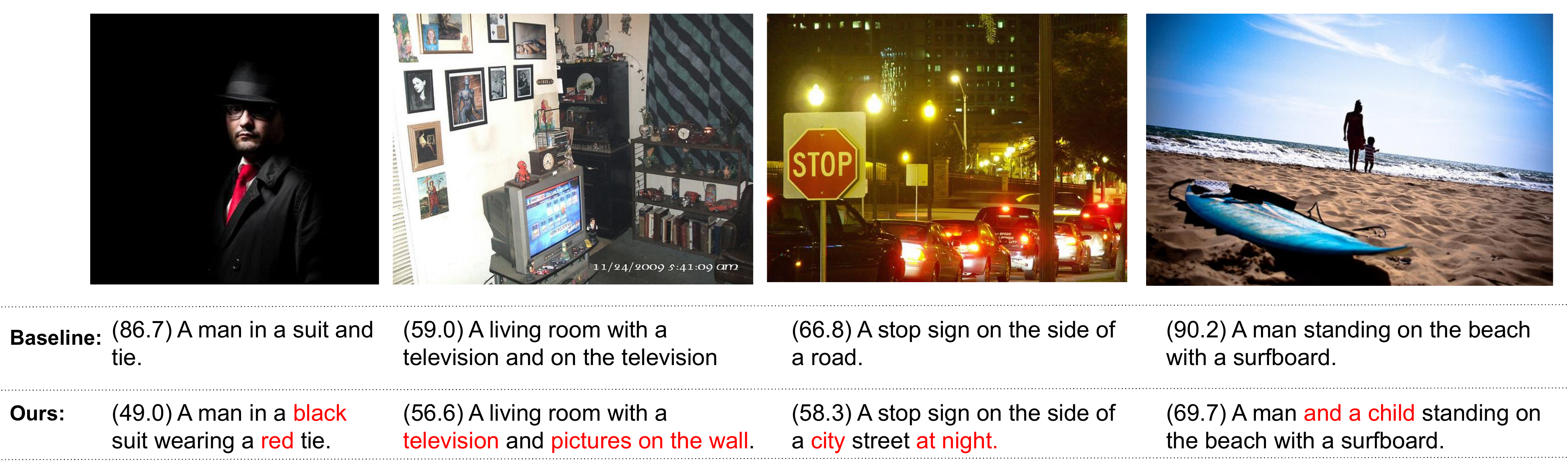}
		\caption{Example captions from the baseline model and our model. The distinctive words are highlighted. The number in parenthesis is the CIDErBtw score, with lower values meaning more distinctive.
		}
		\label{fig:qualitative_1}
\end{figure}

We next show qualitative results for the baseline model Transformer+SCST, and our model Transformer+SCST+CIDErBtw in Figure~\ref{fig:qualitative_1}.
The baseline model generates captions that accurately describe the main object, but are quite generic and monotonous. Intuitively, in order to increase a caption's distinctiveness, the model should focus on more properties that would distinguish the image from others, such as color, numbers, or other objects/background in the image. Our method focuses on more of these aspects and generates accurate results.
% while captions generated by our model are more distinctive in the following aspects. 
Our captions describe more properties of the main object, such as \quotecap{black suit}, \quotecap{red tie} and \quotecap{a man and a child}. We also describe backgrounds that are distinctive, such as \quotecap{pictures on the wall} and \quotecap{city street at night}.

\begin{figure}[tb]
        \centering
        \includegraphics[width=12cm]{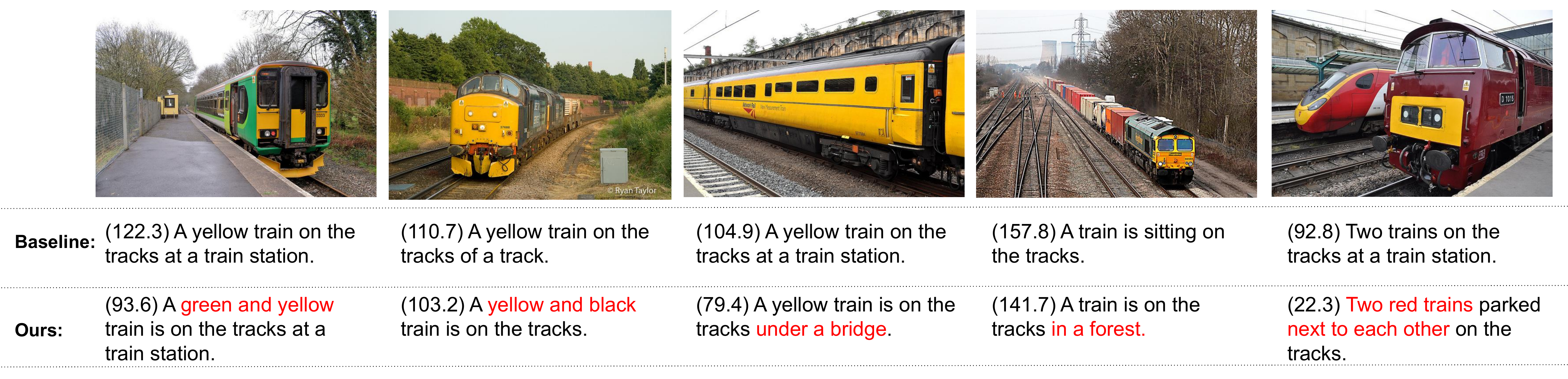}
        \caption{Example captions for a set of similar images.
        }
        \label{fig:qualitative_2} 
\end{figure}

In order to show the distinctiveness of our model, we present a  similar images set with the same semantic meaning in Figure~\ref{fig:qualitative_2}.
The baseline model generates captions that follow generic templates, e.g.  \quotecap{train on the track} or \quotecap{at a train station}. Although the captions are correct, it is hard to tell the images apart according to the captions. Our model enriches the description by mentioning the colors, e.g. the \quotecap{green and yellow} and \quotecap{yellow and black} distinguishing the first two images, and the background environment, e.g. \quotecap{under a bridge} and \quotecap{in a forest}. Furthermore, our model is more sensitive to the relative positions of objects, e.g. \quotecap{next to each other on the tracks}. However, a more descriptive caption may also lead to some errors. For instance, the train in the third image is not actually \quotecap{under a bridge}. More details are in the supplementary material.

\section{Conclusion}

In this paper, we consider 
an important property, \textit{distinctiveness} of image captions,  
and proposed a metric CIDErBtw to evaluate distinctiveness, which can be calculated quickly and easily implemented. We found that human annotations for each image vary in distinctiveness based on CIDErBtw.  To improve the distinctiveness of generated captions, we developed a novel training strategy, where each human ground-truth annotation is assigned a weight based on its distinctiveness computed by CIDErBtw.  Thus, during training the model pays more attention to the captions that are more distinctive. 
We also consider using CIDErBtw directly as part of the reward in RL. 
Extensive experiments were conducted, and we showed that our method is widely applicable to many captioning models. 
Experimental results demonstrate that our training strategy is able to improve both accuracy and distinctiveness, achieving state-of-the-art performance on CIDEr, CIDErBtw and retrieval metrics (R@$K$).

\subsubsection{Acknowledgments.}
This work was supported by grants from the Research Grants Council of the Hong Kong Special Administrative Region, China (Project No. CityU 11212518) and from City University of Hong Kong (Strategic Research Grant No. 7005218).

\clearpage
\bibliographystyle{splncs04}
\bibliography{mainbib}

\begin{thebibliography}{10}
\providecommand{\url}[1]{\texttt{#1}}
\providecommand{\urlprefix}{URL }
\providecommand{\doi}[1]{https://doi.org/#1}

\bibitem{spice}
Anderson, P., Fernando, B., Johnson, M., Gould, S.: {SPICE}: Semantic
  propositional image caption evaluation. In: ECCV (2016)

\bibitem{updown}
Anderson, P., He, X., Buehler, C., Teney, D., Johnson, M., Gould, S., Zhang,
  L.: Bottom-up and top-down attention for image captioning and visual question
  answering. In: CVPR. pp. 6077--6086 (2018)

\bibitem{convimagecap}
Aneja, J., Deshpande, A., Schwing, A.: Convolutional image captioning. In: CVPR
  (2018)

\bibitem{chunseong2017attend}
Chunseong~Park, C., Kim, B., Kim, G.: Attend to you: Personalized image
  captioning with context sequence memory networks. In: CVPR. pp. 895--903
  (2017)

\bibitem{cgan}
Dai, B., Fidler, S., Urtasun, R., Lin, D.: Towards diverse and natural image
  descriptions via a {Conditional GAN}. In: ICCV (2017)

\bibitem{contrastive}
Dai, B., Lin, D.: Contrastive learning for image captioning. In: NeurIPS (2017)

\bibitem{M}
Denkowski, M., Lavie, A.: Meteor universal: Language specific translation
  evaluation for any target language. In: EACL Workshop (2014)

\bibitem{posg}
Deshpande, A., Aneja, J., Wang, L., Schwing, A.G., Forsyth, D.: Fast, diverse
  and accurate image captioning guided by part-of-speech. In: CVPR (2019)

\bibitem{devlin2018bert}
Devlin, J., Chang, M.W., Lee, K., Toutanova, K.: {BERT}: Pre-training of deep
  bidirectional transformers for language understanding. NAACL  (2018)

\bibitem{faghri2017vse++}
Faghri, F., Fleet, D.J., Kiros, J.R., Fidler, S.: {VSE++}: Improving
  visual-semantic embeddings with hard negatives. BMVC  (2018)

\bibitem{templatemodel}
Farhadi, A., Hejrati, M., Sadeghi, M.A., Young, P., Rashtchian, C.,
  Hockenmaier, J., Forsyth, D.: Every picture tells a story: Generating
  sentences from images. In: ECCV (2010)

\bibitem{gu2018stack}
Gu, J., Cai, J., Wang, G., Chen, T.: Stack-captioning: Coarse-to-fine learning
  for image captioning. In: AAAI (2018)

\bibitem{he2017mask}
He, K., Gkioxari, G., Doll{\'a}r, P., Girshick, R.: Mask r-cnn. In: CVPR. pp.
  2961--2969 (2017)

\bibitem{he2016deep}
He, K., Zhang, X., Ren, S., Sun, J.: Deep residual learning for image
  recognition. In: CVPR. pp. 770--778 (2016)

\bibitem{lstm1997}
Hochreiter, S., Schmidhuber, J.: Long short-term memory. Neural computation
  \textbf{9}(8),  1735--1780 (1997)

\bibitem{huang2019attention}
Huang, L., Wang, W., Chen, J., Wei, X.Y.: Attention on attention for image
  captioning. In: ICCV. pp. 4634--4643 (2019)

\bibitem{karpathy2015deep}
Karpathy, A., Fei-Fei, L.: Deep visual-semantic alignments for generating image
  descriptions. In: CVPR. pp. 3128--3137 (2015)

\bibitem{kingma2014adam}
Kingma, D.P., Ba, J.: Adam: A method for stochastic optimization. In: ICLR
  (2015)

\bibitem{R}
Lin, C.Y.: Rouge: A package for automatic evaluation of summaries. In: ACL
  Workshop (2004)

\bibitem{mscocodataset}
Lin, T.Y., Maire, M., Belongie, S., Hays, J., Perona, P., Ramanan, D.,
  Doll{\'a}r, P., Zitnick, C.L.: {Microsoft COCO}: Common objects in context.
  In: ECCV (2014)

\bibitem{liu2019generating}
Liu, L., Tang, J., Wan, X., Guo, Z.: Generating diverse and descriptive image
  captions using visual paraphrases. In: CVPR. pp. 4240--4249 (2019)

\bibitem{liu2018show}
Liu, X., Li, H., Shao, J., Chen, D., Wang, X.: Show, tell and discriminate:
  Image captioning by self-retrieval with partially labeled data. In: ECCV. pp.
  338--354 (2018)

\bibitem{luo2018discriminability}
Luo, R., Price, B., Cohen, S., Shakhnarovich, G.: Discriminability objective
  for training descriptive captions. In: CVPR. pp. 6964--6974 (2018)

\bibitem{mrnn}
Mao, J., Xu, W., Yang, Y., Wang, J., Huang, Z., Yuille, A.: Deep captioning
  with multimodal recurrent neural networks (m-rnn). In: ICLR (2015)

\bibitem{bleu}
Papineni, K., Roukos, S., Ward, T., Zhu, W.J.: {BLEU}: a method for automatic
  evaluation of machine translation. In: ACL (2002)

\bibitem{attend2u:2018:TPAMI}
Park, C.C., Kim, B., Kim, G.: {Towards Personalized Image Captioning via
  Multimodal Memory Networks}. In: IEEE TPAMI (2018)

\bibitem{radford2019language}
Radford, A., Wu, J., Child, R., Luan, D., Amodei, D., Sutskever, I.: Language
  models are unsupervised multitask learners. OpenAI Blog  \textbf{1}(8), ~9
  (2019)

\bibitem{ren2015faster}
Ren, S., He, K., Girshick, R., Sun, J.: {Faster R-CNN}: Towards real-time
  object detection with region proposal networks. In: NeurIPS. pp. 91--99
  (2015)

\bibitem{rennie2017self}
Rennie, S.J., Marcheret, E., Mroueh, Y., Ross, J., Goel, V.: Self-critical
  sequence training for image captioning. In: CVPR. pp. 7008--7024 (2017)

\bibitem{cgan1}
Shetty, R., Rohrbach, M., Hendricks, L.A.: Speaking the same language: Matching
  machine to human captions by adversarial training. In: ICCV (2017)

\bibitem{vggnet}
Simonyan, K., Zisserman, A.: Very deep convolutional networks for large-scale
  image recognition. In: ICLR (2015)

\bibitem{divmetric2}
Van~Miltenburg, E., Elliott, D., Vossen, P.: Measuring the diversity of
  automatic image descriptions. In: COLING. pp. 1730--1741 (2018)

\bibitem{vaswani2017attention}
Vaswani, A., Shazeer, N., Parmar, N., Uszkoreit, J., Jones, L., Gomez, A.N.,
  Kaiser, {\L}., Polosukhin, I.: Attention is all you need. In: NeurIPS. pp.
  5998--6008 (2017)

\bibitem{C}
Vedantam, R., Lawrence~Zitnick, C., Parikh, D.: {CIDEr}: Consensus-based image
  description evaluation. In: CVPR. pp. 4566--4575 (2015)

\bibitem{vered2019joint}
Vered, G., Oren, G., Atzmon, Y., Chechik, G.: Joint optimization for
  cooperative image captioning. In: CVPR. pp. 8898--8907 (2019)

\bibitem{NIC}
Vinyals, O., Toshev, A., Bengio, S., Erhan, D.: Show and tell: A neural image
  caption generator. In: CVPR (2015)

\bibitem{cnnpluscnn}
Wang, Q., Chan, A.B.: {CNN+CNN}: Convolutional decoders for image captioning.
  CVPR Workshop  (2018)

\bibitem{my-div-paper}
Wang, Q., Chan, A.B.: Describing like humans: on diversity in image captioning.
  In: CVPR (2019)

\bibitem{wang2019towards}
Wang, Q., Chan, A.B.: Towards diverse and accurate image captions via
  reinforcing determinantal point process. arXiv  (2019)

\bibitem{Dist-K}
Xu, J., Ren, X., Lin, J., Sun, X.: {Diversity-promoting GAN}: A cross-entropy
  based generative adversarial network for diversified text generation. In:
  EMNLP. pp. 3940--3949 (2018)

\bibitem{spatt}
Xu, K., Ba, J., Kiros, R., Cho, K., Courville, A., Salakhudinov, R., Zemel, R.,
  Bengio, Y.: Show, attend and tell: Neural image caption generation with
  visual attention. In: ICML (2015)

\bibitem{yao2019hierarchy}
Yao, T., Pan, Y., Li, Y., Mei, T.: Hierarchy parsing for image captioning. In:
  ICCV. pp. 2621--2629 (2019)

\end{thebibliography}

\end{document}